\documentclass[journal]{IEEEtai}

\usepackage[colorlinks,urlcolor=blue,linkcolor=blue,citecolor=blue]{hyperref}

\usepackage{color,array}

\usepackage{graphicx}

\usepackage{xcolor}
\usepackage{tabularx} 
\usepackage{booktabs}  
\usepackage{hyperref}
\usepackage{algorithm}
\usepackage{algorithmic}
\usepackage{amssymb}
\usepackage{amsmath}
\usepackage{xcolor}
\usepackage{multirow} 
\usepackage{bbding}
\usepackage{booktabs}
\usepackage{subfig}

\begin{document}

\title{Air Quality Prediction with A Meteorology-Guided Modality-Decoupled Spatio-Temporal Network}

\author{Hang Yin, Yan-Ming Zhang, Jian Xu, Jian-Long Chang, Yin Li, Cheng-Lin Liu, \IEEEmembership{Fellow, IEEE}
\thanks{This work was supported by Beijing Municipal Science and Technology Project (No.Z231100010323005). (Corresponding author: Cheng-Lin Liu.)}

\thanks{Hang Yin, Yan-Ming Zhang, Jian Xu and Cheng-Lin Liu are with the School of Artificial Intelligence, University of Chinese Academy of Sciences, Beijing 100049, China, and also with the State Key Laboratory of Multimodal Artificial Intelligence Systems (MAIS), Institution of Automation, Chinese Academy of Sciences, Beijing 100190, China (e-mail: yinhang2021@ia.ac.cn, ymzhang@nlpr.ia.ac.cn, jian.xu@ia.ac.cn, liucl@nlpr.ia.ac.cn).}
\thanks{Jian-Long Chang and Yin Li are with Huawei Cloud Enterprise Intelligence, BeiJing, China, 100085 (e-mail: jianlong.chang@huawei.com, liyin9@huawei.com).}

}

\maketitle

\begin{abstract}
Air quality prediction plays a crucial role in public health and environmental protection. Accurate air quality prediction is a complex multivariate spatiotemporal problem, that involves interactions across temporal patterns, pollutant correlations, spatial station dependencies, and particularly meteorological influences that govern pollutant dispersion and chemical transformations. Existing works underestimate the critical role of atmospheric conditions in air quality prediction and neglect comprehensive meteorological data utilization, thereby impairing the modeling of dynamic interdependencies between air quality and meteorological data. To overcome this, we propose MDSTNet, an encoder-decoder framework that explicitly models air quality observations and atmospheric conditions as distinct modalities, integrating multi-pressure-level meteorological data and weather forecasts to capture atmosphere-pollution dependencies for prediction. Meantime, we construct ChinaAirNet, the first nationwide dataset combining air quality records with multi-pressure-level meteorological observations. Experimental results on ChinaAirNet demonstrate MDSTNet's superiority, substantially reducing 48-hour prediction errors by  17.54\% compared to the state-of-the-art model. The source code and dataset will be available on github.
\end{abstract}

\begin{IEEEImpStatement}
Air quality prediction serves as a critical tool for preventive public health management by enabling early warnings to reduce pollution-induced disease burdens. As a multivariate spatiotemporal forecasting task, it requires modeling multi-location pollutant concentrations through time. Existing predictive models mostly underestimate the critical influence of meteorological conditions. This paper proposes the Meteorology-Guided Modality-Decoupled Spatio-Temporal Network (MDSTNet). MDSTNet decouples air quality observations and meteorological conditions as two distinct modalities while integrating weather forecast as dynamic input prompts to guide air quality prediction. MDSTNet is supposed to effectively reduce prediction errors, establishing a new paradigm for meteorology-guided air quality prediction. Meanwhile, we also built a nationwide dataset named ChinaAirNet, which contains two years air quality observations integrated with multi-pressure level meteorological data. As an open-access resource, ChinaAirNet enhances research reproducibility while providing the first unified benchmark for systematically modeling multi-pressure level meteorological factors and air pollution interactions.

\end{IEEEImpStatement}


\section{Introduction}

\IEEEPARstart{T}{he} urbanization and industrialization of the world have inevitably led to serious air pollution problems. Exposure to air pollution is directly linked to the prevalence of non-communicable diseases such as heart disease, asthma, and lung cancer~\cite{WHO, matus_health_2012}, and the high cost of treatment places a massive burden on governments and individuals. This situation can be greatly mitigated if appropriate countermeasures, such as reminding people to wear masks, limiting factory emissions, and restricting vehicles, are taken in advance. Therefore, accurate air quality prediction has become a research hotspot for public health, national economy, and urban management.


Air quality prediction as a multivariate spatiotemporal prediction problem involves jointly forecasting the concentrations of various pollutants at multiple locations for a specified time range. The COVID-19 lockdowns demonstrated that even with drastic anthropogenic emission reductions, air quality improvements remained limited. This indicated that air quality variations were predominantly influenced by meteorological conditions~\cite{COVID}. Studies~\cite{Meteorology_aq_beijing,MeteEffect, LLJ} highlight that the interactions between meteorological factors and air pollution exhibit complex temporal and spatial dynamics, with nonlinear effects playing a critical role in pollutant formation and dispersion. Therefore, the main difficulty of accurate air quality prediction lies in comprehensive modeling of the complex spatiotemporal correlations between meteorological factors and air pollution, which primarily stems from two challenges. First, air quality is characterized by pollutant dispersion patterns while weather follows pressure-driven fluid dynamics, which exhibit distinct spatiotemporal patterns. Therefore, air quality and meteorological conditions constitute distinct modalities that require separate modeling to prevent cross-modal interference. Second, AI weather models~\cite{aifsecmwf,graphcast, gencast, pangu} provide high-accuracy weather forecasts that could significantly enhance air quality prediction. The temporal gap between weather forecasts and historical air quality observations complicates their integration.

Despite the numerous efforts in advancing air quality prediction research in recent years~\cite{GC-DCRNN,Deepair,PM2.5-GNN,liang2023airformer, STAFNet,han2022semi,airphynet}, existing approaches predominantly frame this task as a multivariate spatiotemporal prediction problem. These conventional methods~\cite{DALDeepAirLearning,GAGNN} focus on modeling either the temporal dynamics inherent in air quality data or the spatial dependencies between monitoring stations, while frequently underestimating the critical influence of meteorological conditions that fundamentally govern pollutant dispersion and chemical transformations. On the one hand, all these studies~\cite{GC-DCRNN,Deepair,PM2.5-GNN} are confined to utilizing only historical air quality observations or integrating limited surface-level meteorological features for forecasting, while disregarding the predictive value of upper-air pressure levels meteorological data and weather forecasting data.
On the other hand, conventional approaches~\cite{liang2023airformer, han2022semi,airphynet} often naively concatenate air quality with meteorological features as homogeneous input vectors. However, air quality exhibit starkly dissimilar spatiotemporal patterns compared to meteorological fluid dynamics. Such naive fusion forces both modalities into a unified spatiotemporal dependency and fundamentally misrepresents their heterogeneous modalities, impairing the model's capacity to capture cross-modal dependencies and meteorology-guided pollutant dispersion processes.

To overcome the above shortcomings, in this paper, we propose Meteorology-Guided \textbf{M}odality-\textbf{D}ecoupled \textbf{S}patio-\textbf{T}emporal \textbf{Net}work (MDSTNet), a novel deep learning framework that explicitly and carefully models meteorological conditions for air quality prediction through cross-modal interaction. MDSTNet employs an encoder-decoder framework, where the encoder formalizes air quality observations and meteorological conditions as two distinct modalities. The decoder adaptively aligns weather forecast with historical air quality representations to resolve temporal asynchrony. Specifically, the encoder integrates stacked transformer layers, with each layer consisting three decoupled parallel attention branches to comprehensively capture historical patterns from three views: spatial correlations within air quality monitoring stations, inter-variable dependencies among diverse pollutants, and cross-modal interactions between pollutants and meteorological features. The decoder utilizes multi-step future weather forecast data as contextual prompts to dynamically filter these learned historical representations through attention mechanisms in spatial and variable dimensions, ultimately generating predictive representations for accurate air quality forecasting.

Along with the proposed architecture, we also built a new large-scale dataset named ChinaAirNet, which contains two years of nationwide air quality observations from 1,628 monitoring stations in China. Distinct from existing datasets, ChinaAirNet pioneers the comprehensive integration of air quality records with multi-pressure-level meteorological data (56 meteorological variables from 1000 hPa to 850 hPa), aligning with air quality data at location and time. 
 
Our research contributions can be summarized as follows.
\begin{itemize}
\item[$\bullet$] We present the first systematic incorporation of multi-pressure-level meteorological data and weather forecast information into air quality prediction, empirically unveiling the critical role of meteorological  conditions.

\item[$\bullet$] We formalize air quality observations and meteorological conditions as two distinct modalities and design a novel multi-modal encoder-decoder architecture named MDSTNet, where the encoder models different types of dependencies with decoupled parallel branches and an efficient cross-attention technique, while the decoder incorporates weather forecast features as dynamic input prompts to guide air quality prediction.

\item[$\bullet$] We construct the first large-scale open-source air quality dataset, ChinaAirNet, which integrates air quality records and multi-pressure-level meteorological observations.  

\item[$\bullet$] Extensive experimental results demonstrate the effectiveness of our method. Specifically, MDSTNet achieves 17.54\% improvement in 48-hour prediction compared to the state-of-the-art model on ChinaAirNet.
\end{itemize}

\section{Related Work}

\subsection{Air Quality Prediction}

During the past decades, air quality prediction methodologies have evolved through three principal paradigms: physics-based models numerical models, statistical approaches and deep learning models. Numerical models such as Early conventional methods~\cite{vardoulakis2003modelling,arystanbekova2004application,daly2007air} exploit domain knowledge to simulate the pollutant dispersion through physicochemical mechanisms like atmospheric dynamics and chemical reactions. However, their reliance on accurate source data and prohibitive computational demands significantly constrained practical deployment. Statistical approaches like LR~\cite{donnelly2015real} and ARIMA~\cite{kumar2010arima} captured basic temporal patterns but inadequately modeled complex spatiotemporal interactions and meteorological influences. 

Recently, the accumulation of historical air quality data has driven the prominence of deep learning approaches in air quality forecasting. GC-DCRNN~\cite{GC-DCRNN} construct a geo-context based graph and integrates recurrent neural networks with diffusion convolutions to to handle the spatial correlations and temporal dependencies. While DeepAir~\cite{Deepair} utilizes CNN and Bi-LSTM to extract local spatial features and spatial-temporal dependencies separately, and AirFormer~\cite{liang2023airformer} introduced variant transformers to dynamically model spatiotemporal information. AirPhyNet~\cite{airphynet} incorporates physics processes into a deep learning framework to govern air pollutant transport. These models predominantly rely on historical pollutant measurements and limited surface-level meteorological data, neglecting multi-pressure-level meteorological variables and weather forecast data.

In terms of model architecture, SSH-GNN~\cite{han2022semi} employs hierarchical graph neural networks, which can encode large-range spatiotemporal dependencies with various granularities by constructing a multi-level hierarchy. Those methods naively concatenate meteorological or weather forecast features with air quality features as network input and share the same dependency patterns in all variates channels, which can be detrimental if the underlying multiple variates series exhibit different spatiotemporal behaviors according to PatchTST~\cite{patchtst}.
MasterGNN~\cite{han2021joint} adopts a heterogeneous graph model the spatial and temporal correlations between air quality stations and meteorological stations. While STAFNet~\cite{STAFNet} employs dual independent graphs with cross-graph attention to resolve interaction, both approaches fail to operationalize future weather forecasts as dynamic constraints for air quality prediction. PM2.5-GNN~\cite{PM2.5-GNN} integrating GNN and gated recurrent units (GRU) to predict PM\textsubscript{2.5} concentrations by explicitly encoding meteorological information and weather forecast into nodes and edges attribute. CGF~\cite{CGF} enhances prediction accuracy by incorporating PM\textsubscript{2.5} category information and treating weather forecasts as auxiliary future features within the decoder. However, both models adopt a single-step forecasting paradigm, restricting their capacity to explicitly model long-term temporal dependencies and limiting the utilization of multi-step weather forecasts. 

While learning-based models generally achieve superior prediction performance compared to numerical and statistical counterparts due to the representational capacity of deep neural networks, existing approaches inadequately capture the complex interdependencies among air quality patterns, meteorological characteristics, and multi-step weather forecasts. Unlike prior works that jointly input air quality and meteorological data into unified networks, our framework treats these features as distinct modalities for independent modeling while fully leveraging comprehensive meteorological data across multiple pressure levels and multi-horizon weather forecasts.

\begin{figure*}[htp]
\centering
\includegraphics[width=0.95\textwidth]{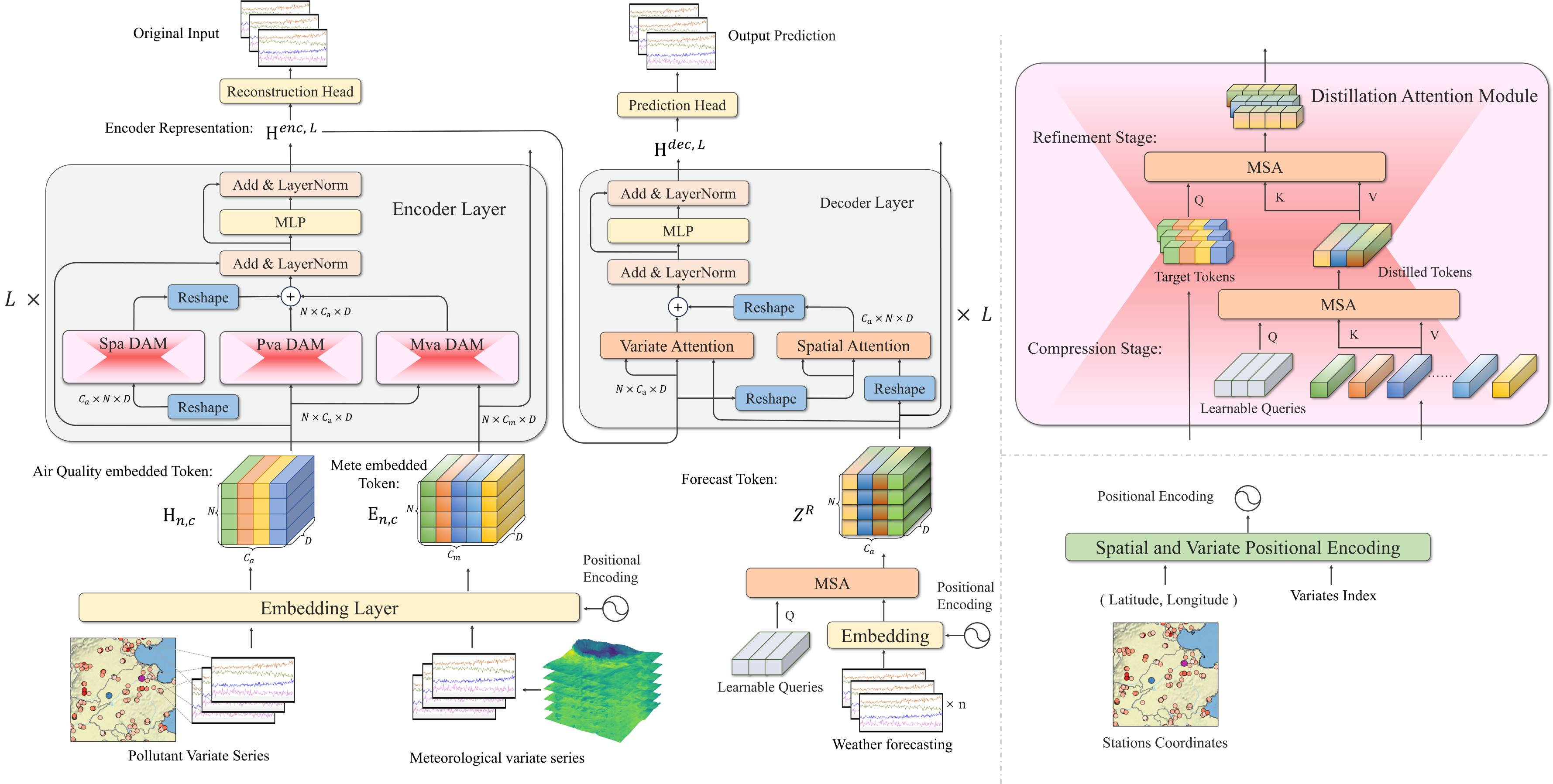} 
\caption{ Architectural Overview of MDSTNet. The framework employs an encoder-decoder structure, with each encoder layer comprising three parallel branches: Spatial Attention (Spa), Pollutant Variate Attention (Pva), and Meteorological Variate Attention (Mva). Each branch is implemented as an independent Distillation Attention Module (DAM). The decoder is also composed of a multi-layer transformer structure. The structure of DAM is illustrated in the upper right corner, while the lower right inset demonstrates the integration of Spatial and Variate Positional Encoding.}
\label{framework}
\end{figure*}

\section{Method}
Our goal is to forecast the future concentrations of major pollutants based on historical observations and meteorological variates (e.g., historical meteorological data and weather forecasts) gathered from $N$ monitoring stations. The air quality observations of station $n$ at time $t$ are denoted as $\mathbf{x}_n^t\in \mathbb {R}^{C_{a}}$, which is a vector of air pollutant concentrations.  
Each station is associated with meteorological variates $\mathbf{y}_n^t\in \mathbb {R}^{C_m}$, aligned in both spatial location and time stamp. Similarly, we represent weather forecasting variates as $\hat{\mathbf{y}}_n^t\in \mathbb {R}^{C_{m}}$. The $C_{a},{C_m}$ represent the number of pollutant and meteorological variates, respectively. In addition, we denote $\mathbf{X}^{t}=\left\{\mathbf{x}_1^{t}, \mathbf{x}_2^{t}, \ldots, \mathbf{x}_N^{t}\right\}\in \mathbb {R}^{ N \times C_{a}}$, 
$\mathbf{Y}^{t}=\left\{\mathbf{y}_1^{t}, \mathbf{y}_2^{t}, \ldots, \mathbf{y}_N^{t}\right\}\in \mathbb {R}^{  N \times C_m }$, 
$\hat{\mathbf{Y}}^{t}=\left\{\hat{\mathbf{y}}_1^{t}, \hat{\mathbf{y}}_2^{t}, \ldots, \hat{\mathbf{y}}_N^{t}\right\}\in \mathbb {R}^{  N \times C_m }$.
Thus, the air quality prediction problem is defined as predicting the concentrations over the next $\tau$ time steps:
\begin{equation}
 \mathcal{F}(\mathbf{X}^{1:T},\mathbf{Y}^{1:T},\hat{\mathbf{Y}}^{T+1:T+\tau}) \rightarrow \hat{\mathbf{X}}^{T+1:T+\tau},
\end{equation}
where $\mathbf{X}^{1:T}\in \mathbb {R}^{  N \times C_{a} \times T }$
represents the historical pollutant concentrations for $N$ locations from time step $1$ to $T$, $\mathbf{Y}^{1:T} \in \mathbb {R}^{ N \times C_{m} \times T }$ and $\hat{\mathbf{Y}}^{T+1:T+\tau} \in \mathbb {R}^{ N \times C_{m} \times \tau }$ represents the meteorological variates and weather forecasts. 
$\hat{\mathbf{X}}^{T+1:T+\tau} \in \mathbb {R}^{ N \times C_{a} \times \tau } $ is the predicted concentrations of all the stations in future $\tau$ time steps, and $\mathcal{F}(\cdot)$ denotes the mapping function we aim to learn.

As illustrated in Fig.~\ref{framework}, our proposed MDSTNet adopts an architecture comprising three core components: an embedding layer, an encoder-decoder backbone, and dual-task heads for reconstruction and prediction. The embedding layer is responsible for tokenizing the time series to capture temporal dynamics and incorporating spatial information through positional encoding. The encoder architecture consists of sequentially stacked transformer layers, with decoupled parallel attention branches within each layer, designed to model the multiple correlations in historical air quality and meteorological data. The decoder, structured with analogous depth,  leverages future weather forecasts as conditional prompts to dynamically align comprehensive historical representations from the encoder, establishing spatiotemporal dependencies critical for multi-step prediction tasks.

\subsection{Embedding Layer}

\subsubsection{Series Embedding}
Inspired by iTransformer~\cite{liu2024itransformer}, the series of different stations and different variates are embedded independently into a $D$-dimensional token with a multi-layer perceptron (MLP): $\mathbb{R}^{T} \rightarrow \mathbb{R}^{D}$. Notably, weather forecast sequences with heterogeneous temporal lengths require distinct MLPs for series embedding. We deliberately omit instance normalization, a common practice in time-series forecasting, to preserve absolute magnitude information that carries critical physical implications for air quality prediction.

\subsubsection{Spatial and Variate Positional Encoding}
Due to the permutationally invariant property, transformers are position-insensitive. 
Thus, positional encoding is needed to enable Transformer to process spatiotemporal data. In contrast to the discrete positions utilized in NLP/CV~\cite{transformer,rope}, the spatial locations are continuous values. To achieve that, similar to the positional encoding in vanilla Transformer~\cite{transformer}, we design a simple and effective 2D position encoding:
\begin{equation}
\begin{cases}
 P E_{((\phi_n, \lambda_n), 4 i)} & =\sin \left(\phi_n/ 10000^{2 i /D     }\right)\\
 P E_{((\phi_n, \lambda_n), 4 i+2)} & =\cos \left(\phi_n / 10000^{2 i / D}\right)\\
 P E_{((\phi_n, \lambda_n), 4 i+1)} & =\sin \left(\lambda_n / 10000^{2 i / D}\right)\\
P E_{((\phi_n, \lambda_n), 4 i+3)} & =\cos \left(\lambda_n/ 10000^{2 i / D}\right),
\end{cases}
\end{equation}
where $PE\in \mathbb {R}^{N \times D }$, $(\phi_n, \lambda_n)$ is the latitude and longitude of station $n$,  $i$ is the dimension and $D$ is the hidden dimension. 
Similarly, we incorporate vanilla positional encoding in the channel dimension: $CE\in \mathbb {R}^{C \times D }$ to facilitate the capture of correlations among variates. The embedding process can be formally expressed as follows:
\begin{equation}
\begin{aligned}
\mathbf{H}_{n,c} & =\operatorname{MLP}\left(\mathbf{X}_{n,c}^{1:T}\right) + PE_(\phi_n, \lambda_n) +CE_c, \\
\mathbf{E}_{n,c} & =\operatorname{MLP}\left(\mathbf{Y}_{n,c}^{1:T}\right) + PE_(\phi_n, \lambda_n) +CE_c, \\
\mathbf{Z}_{n,c} & =\operatorname{MLP}\left(\hat{\mathbf{Y}}_{n,c}^{T+1:T+\tau}\right) + PE_(\phi_n, \lambda_n) +CE_c, \\
\end{aligned}
\end{equation}
where $\mathbf{H}_{n,c}$ and $ \mathbf{E}_{n,c}$ is D-dimension embedded hidden tokens of historical air quality and meteorological variates, while $ \mathbf{Z}_{n,c}$ represents
 embeddings for future weather forecast variables.

\subsection{Encoder}

The encoder consists of $L$ stacked identical layers. As illustrated in Fig.~\ref{framework}, the encoder is organized in a parallel structure to exploit spatial relationships, pollutant correlations, and exogenous meteorological support with three decoupled attention branches. For the $l$-th encoder layer, the input is the embedded pollutant variate tokens: $\mathbf{H}^{l-1} \in \mathbb{R}^{N \times C_a \times D}$ and the embedded meteorological variate tokens: $\mathbf{E}  \in \mathbb{R}^{N \times C_m \times D}$. The process with residual connections can be formulated as follows:
\begin{align}
 \mathbf{H}^{enc,l}  &= \operatorname{Encoder}\left(\mathbf{H}^{enc,l-1}, \mathbf{E}^{} \right) +\mathbf{H}^{enc,l-1} &l =1,\dots,L. 
\end{align}

\subsubsection{Spatial Attention Branch}
To capture the spatial correlations among the $N$ stations, Multi-head Self-Attention (MSA) is applied across the $N$ tokens $\mathbf{H}_{:,c}^{l-1}\in \mathbb{R}^{N\times D}$ for each pollutant variate $c$. We use $\mathbf{H}_{}^\text{spa}$ to represent the inter-station correlations modeled through spatial attention (Spa) as follows:
\begin{align}
\mathbf{H}_{:,c}^\text{Spa} &=  \operatorname{Attention}\left(\mathbf{H}_{:,c}^{enc,l-1},\; \mathbf{H}_{:,c}^{enc,l-1}\right)  &c =1,\dots,C_a. \label{spa}
\end{align}

\subsubsection{Pollutant Variate Attention Branch}
For the embedded pollutant variates token in stations $n$: $\mathbf{H}_{n,:}^{l-1}\in \mathbb{R}^{C_a\times D}$, we apply pollutant variate attention (Pva) for pollutant variate correlations $\mathbf{H}_{}^\text{Pva}$, as follows:
\begin{align}
\mathbf{H}_{n,:}^\text{Pva} &=  \operatorname{Attention}\left(\mathbf{H}_{n,:}^{enc,l-1},\; \mathbf{H}_{n,:}^{enc,l-1}\right) &n =1,\dots,N. \label{pva}
\end{align}

\subsubsection{Meteorological Variate Attention Branch} Similarly, this Branch is applied to processes meteorological variates. In contrast, the dimensions of meteorological variates $C_m$ differ from  $C_a$. Therefore, we use cross-attention (Mva) to exploit the exogenous meteorological support $\mathbf{H}_{n,:}^\text{Mva}$ of the embedded meteorological variates token in stations $n$: $\mathbf{E}_{n,:}\in \mathbb{R}^{C_m\times D}$, as follows:
\begin{align}
\mathbf{H}_{n,:}^\text{Mva} &=  \operatorname{Attention}\left(\mathbf{H}_{n,:}^{enc,l-1},\; \mathbf{E}_{n,:}\right) &n =1,\dots,N. \label{mva}
\end{align}

In Equation~\ref{spa}--\ref{mva}, $\operatorname{Attention}
(\mathbf{A},\mathbf{B})$ is the attention operation with $\mathbf{A}$ as queries and $\mathbf{B}$ as keys and values. Finally, encoder outputs the updated pollutant variate tokens as:
\begin{equation}
    \operatorname{Encoder}(\mathbf{H}_{enc}^{l-1}, \mathbf{E}) = \mathbf{H}^\text{Spa} + \mathbf{H}^\text{Pva} + \mathbf{H}^\text{Mva}.
\end{equation}

However, the computational cost of the classic attention is $\mathcal {O}(MN)$ due to the calculation of the similarity among $M$ queries and $N$ keys. This computational load becomes prohibitive when handling data with a large number of tokens. To address these issues, we introduced Distillation Attention Module (DAM) which aims to reduce calculations and improve information quality as described in section~\ref{sec:DAM}.

\subsection{Distillation Attention Module}\label{sec:DAM}
As presented in Fig.~\ref{framework}, DAM can be decomposed into two sequential stages: Compression Stage and Refinement Stage. Compression stage compresses the $N$ key/value tokens $\mathbf{B}=\left\{\mathbf{b}_1, \mathbf{b}_2, \ldots, \mathbf{b}_N\right\} \in \mathbb{R}^{ N \times D}$ into a small number of distilled tokens, while Refinement stage obtains the representation $\mathbf{H} $ by aggregating information from the distilled tokens based on the $M$ query tokens $\mathbf{A}=\left\{\mathbf{a}_1, \mathbf{a}_2, \ldots, \mathbf{a}_M\right\} \in \mathbb{R}^{ M \times D}$. 
\subsubsection{Compression Stage}\label{sec:DAM_Compression_Stage}
In this stage, DAM takes $N$ tokens $\mathbf{B} \in \mathbb{R}^{ N \times D}$ as inputs and generates $N_r$ distilled tokens as outputs with $N_r \ll N$. The tokens are first normalized and then mapped to keys and values by a linear layer. Subsequently, $N_r$ learnable vectors are introduced as queries, which are used to distill information from the original $N$ tokens and compress them into $N_r$ tokens. This process can be formulated as:
\begin{align}
\mathbf{Q}_c \in & \mathbb{R}^{ N_r \times D},\quad \mathbf{K}_c = \mathbf{B} \mathbf{W}_c^k,\quad \mathbf{V}_c = \mathbf{B} \mathbf{W}_c^v \\
\mathbf{R}  = & \operatorname{Softmax}\left(\frac{\mathbf{Q}_c \mathbf{K}_c^T }{\sqrt{D}}\right) \mathbf{V}_c,
\end{align}
where $\mathbf{Q}_c\in \mathbb{R}^{ N_r \times D}$ represents $N_r$ learnable queries, $\mathbf{W}_c^k,\mathbf{W}_c^v \in \mathbb{R}^{ D \times D}$ are the weight matrices. $\mathbf{R} \in \mathbb{R}^{ N_r \times D} $ represents the distilled tokens. This stage reduces the number of tokens while removing a substantial amount of redundant information, thereby increasing the signal-to-noise ratio of the tokens.

\subsubsection{Refinement Stage}
Given $M$ target tokens $\mathbf{A} \in \mathbb{R}^{ M \times D}$ and $N_r$ distilled tokens $\mathbf{R}$, Refinement stage obtains the representation $\mathbf{H} $ with the standard cross attention:
\begin{align}
\mathbf{Q}_r =& \mathbf{A} \mathbf{W}_r^q,\quad \mathbf{K}_r = \mathbf{R} \mathbf{W}_r^k,\quad \mathbf{V}_r = \mathbf{R} \mathbf{W}_r^v \\
\mathbf{H}  =& \operatorname{Softmax}\left(\frac{\mathbf{Q}_r \mathbf{K}_r^T }{\sqrt{D}}\right) \mathbf{V}_r,
\end{align}
where $\mathbf{W}_r^k,\mathbf{W}_r^v \in \mathbb{R}^{ D \times D}$ are the weight matrices, $\mathbf{H} \in \mathbb{R}^{ M \times D} $ is the resulting representation.

We use $\operatorname{DAM}(\mathbf{A},\mathbf{B})$ to replace $\operatorname{Attention}(\mathbf{A},\mathbf{B})$ in Equation~\ref{spa},\ref{pva},\ref{mva}. The multi-head attention technique is adopted in both stages. The benefit of DAM is threefold. First, by introducing the distilled tokens, the complexity of DAM is reduced to $\mathcal{O}(NN_r+MN_r)$, which is much lower than $\mathcal{O}(MN)$ when $N_r\ll N$. Second, unlike graph-based models, DAM still enjoys the global receptive field as normal attention which is crucial for long-term and large-range prediction. Third, as we will show in experiments, DAM effectively improves air quality prediction accuracy thanks to the information distillation.

\subsection{Decoder}

As illustrated in Fig.~\ref{framework}, the decoder initially condenses embeddings of future weather forecast variables into $C_a$ compact tokens through the compression stage of DAM, which is described in section~\ref{sec:DAM_Compression_Stage}, effectively aligning the dimensionality with air quality variables while preserving critical forecast information. This process can be formulated as:
\begin{align}
\mathbf{Z}^R_{n,:} &=  \operatorname{Compression Stage}\left(\mathbf{Z} _{n,:}\right) &n =1,\dots,N. 
\end{align}

where $\mathbf{Z}  \in \mathbb{R}^{N \times C_m \times D}$ are the weather forecast embeddings, $\mathbf{Z}^R  \in \mathbb{R}^{N \times C_a \times D}$ is the compact forecast tokens. These compressed tokens serve as dynamic prompts, guiding the decoder's cross-attention operations to establish spatiotemporal dependencies by adaptively correlating future weather conditions with encoded historical patterns of air pollutants and meteorological influences. Specifically, For the $l$-th decoder layer, those forecast prompts are inputted as queries that interact with historical representations $\mathbf{H}$ from the encoder. The process with residual connections can be formulated as follows:
\begin{align}
\mathbf{H}_{n,:}^{dsa, l} &=  \operatorname{Attention}\left(\mathbf{Z}^R_{n,:},\mathbf{H}_{n,:}^{dec, l-1}\; \right)  &n =1,\dots,N. \\
\mathbf{H}_{:,c}^{dva, l} &=  \operatorname{Attention}\left(\mathbf{Z}^R_{:,c},\mathbf{H}_{:,c}^{dec, l-1}\; \right)  &c =1,\dots,C_a.\\
\mathbf{H}_{}^{dec, l} &=  \mathbf{H}_{}^{dsa, l} +\mathbf{H}_{}^{dva, l} +\mathbf{H}_{}^{dec, l-1}
\end{align}
where $\mathbf{H}_{n,:}^{dsa, l} , \mathbf{H}_{:,c}^{dva, l}\in \mathbb{R}^{N \times C_a \times D} $ correspond to dynamic representations in spatial and variable dimensions, respectively. $ \mathbf{H}^{dec, l} \in \mathbb{R}^{N \times C_a \times D}$ denote the output features of the $l$-th decoder layer, and $\mathbf{H}^{dec, l-1}$ serve as keys and values in the attention operation. Notably, for the first decoder layer ($l$=1), the input $\mathbf{H}^{dec, l-1}$ corresponds to the encoder's final output $\mathbf{H}^{enc,L}$. This architecture enables the decoder to condition predictions on both learned historical correlations and weather forecast constraints, effectively addressing the covariate shift between historical and future phases.

\begin{table*}[ht]
\begin{center}
\caption{Datasets description and comparison. Mete Variates indicates the number of meteorological parameters per station. Open Access indicates public availability, with the \XSolidBrush\ symbol denoting non-open datasets.}\label{data_tab}
\begin{tabular}{l | c c c c c c}
\midrule\midrule
Dataset & Scale & \# Stations & Timespan & \# Air Quality Records & \# Mete Variates & Open Access \\ 
\hline   
KDDCUP-Beijing & City & 35 &  2017/1/1 - 2018/3/31 & 1,839,600 & 5 & \Checkmark \\ 
KnowAir & Nation & 184 & 2015/1/1 - 2018/12/31 & 2,150,592 & 17 & \Checkmark \\ 
UrbanAir & Province & 437 & 2018/1/1 - 2020/4/1  & 22,968,720 & 6 & \Checkmark \\ 
SSH-GNN-BTH~\cite{han2022semi} & Province & 479 & 2018/1/1 - 2020/4/1 & 8,544,288 & 5 & \XSolidBrush \\  
AirFormer~\cite{liang2023airformer} & Nation & 1,085 & 2015/1/1 - 2018/12/31 & 222,485,790 & 5 & \XSolidBrush \\ 
\hline
ChinaAirNet & Nation & 1,628 &  2022/1/1 - 2023/12/31 & 199,657,920 & $8\times7$ & \Checkmark \\ 
\midrule\midrule
\end{tabular}
\end{center}
\end{table*}

\subsection{Multi-task Learning}
To enhance the encoder's representation capability in modeling multiple correlations within historical air quality and meteorological data while empowering the decoder to leverage forecast prompts and comprehensive encoded representations for accurate prediction, we implement a joint training framework integrating supervised air quality prediction with self-supervised reconstruction tasks.
\subsubsection{Prediction Task}
As has been proven effective by previous work~\cite{dlinear,liu2024itransformer,TiDE}, we employ a prediction head, which is a MLP to generate future predictions: $\hat{\mathbf{X}}_{n,c}^{T+1:T+\tau} \in \mathbb {R}^{ \tau } $ based on the decoder's output features $\mathbf{H}^{dec, L}$ as follows:
\begin{equation}
\hat{\mathbf{X}}_{n,c}^{T+1:T+\tau} = \operatorname{MLP}\left( \mathbf{H}_{n,c}^{dec, L} \right).
\end{equation}
We compute the Mean Squared Error (MSE) prediction loss between the ground-truth observations and predictions as follows:
\begin{equation}
\mathcal{L}_{prediction}=\frac{1}{N C} \sum_{n=1}^{N} \sum_{c=1}^{C} \big\| \hat{\mathbf{X}}_{n,c}^{T+1:T+\tau} -\mathbf{X}_{n,c}^{T+1:T+\tau}\big\|^2 .
\end{equation}
\subsubsection{Reconstruction Task}
As encoder representation $\mathbf{H}^{enc, L}$, we feed it into the reconstruction head, which shares the same structure as the prediction head, and ultimately reconstruct the original air quality sequence $\mathbf{X}^{1:T}\in \mathbb {R}^{  N \times C_{a} \times T }$ as follows:
\begin{equation}
\hat{\mathbf{X}}_{n,c}^{1:T} = \operatorname{MLP}\left( \mathbf{H}_{n,c}^{enc, L} \right).
\end{equation}
We compute the MSE reconstruction loss for the entire time series as follows:
\begin{equation}
\mathcal{L}_{reconstruction}=\frac{1}{N C} \sum_{n=1}^{N} \sum_{c=1}^{C} \big\| \hat{\mathbf{X}}_{n,c}^{1:T} -\mathbf{X}_{n,c}^{1:T}\big\|^2 .
\end{equation}
MDSTNet is trained through backpropagation by minimizing the overall loss as follows:
\begin{equation}
\mathcal{L}=\mathcal{L}_{prediction}+\mathcal{L}_{reconstruction}.
\end{equation}

\section{Dataset}

While existing air quality prediction datasets such as UrbanAir\footnote{http://urban-computing.com/data/Data-1.zip}, KDDCUP\footnote{https://www.biendata.xyz/competition/kdd\_2018}, and KnowAir\footnote{https://github.com/shuowang-ai/PM2.5-GNN} have advanced research in this field, as presented in Table~\ref{data_tab}, they have severe limitations. Datasets like UrbanAir focus on municipal-scale monitoring from 43 cities, while KDDCUP restricts observations to specific metropolitan clusters, both failing to achieve nationwide spatial coverage. Second, even when incorporating meteorological variables, these datasets predominantly rely on surface-level observations, neglecting essential atmospheric profile variables across vertical pressure levels. In addition, the restricted accessibility of certain proprietary datasets undermines reproducibility and hinders cross-study validation, also constraining their scientific utility. To address these gaps, we present ChinaAirNet — the first nationwide multi-modal environmental dataset integrating air quality monitoring, multi-pressure-level meteorological reanalysis, and Weather forecasts across mainland China. To the best of our knowledge, ChinaAirNet is the largest and the most complete open-source dataset for air quality prediction up to now.
\begin{figure}[t]
    \centering
    \includegraphics[width=1\linewidth]{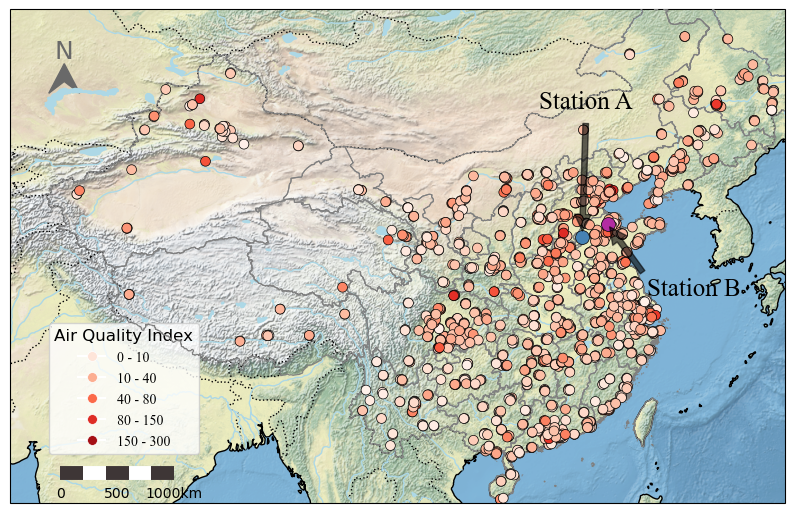}
    \caption{Distribution of air quality monitoring stations in China and corresponding air quality levels.}
    \label{fig:station_distribution}
\end{figure}

\subsection{Collection and Description}
Air quality data: We collected hourly air quality data reported by 2,026 air quality monitoring stations covering 372 cities from National Urban AQ Real-time Release Platform\footnote{http://www.cnemc.cn/en/}. Each air quality observation includes the concentration of six major pollutants (PM\textsubscript{2.5}, PM\textsubscript{10}, O\textsubscript{3}, NO\textsubscript{2}, SO\textsubscript{2}, CO), and derived Air Quality Index (AQI) calculated according to China's MEP-2012 standard. The station selection process prioritized temporal data completeness, retaining 1,628 monitoring stations with a data missing rate of less than 10\%. We use linear interpolation along the temporal dimension to fill in missing values. 

Meteorological data: Following previous work~\cite{metedata}, We utilized the fifth-generation European Centre for Medium-Range Weather Forecasts (ECMWF) atmospheric reanalysis dataset\footnote{https://cds.climate.copernicus.eu/datasets/reanalysis-era5-pressure-levels} (ERA5) with 0.25° latitude-longitude spatial resolution. The selected record encompasses 8 meteorological variables spanning 7 vertical pressure levels from 1000 hPa to 850 hPa, with a vertical resolution of 25 hPa between adjacent levels. These variables include: geopotential height, specific humidity, specific rain water content, air temperature, U and V components of horizontal wind, vertical velocity, and relative vorticity. The dataset was acquired at hourly temporal resolution and subsequently spatially aligned to the geographical coordinates of air quality monitoring stations through bilinear interpolation based on latitude-longitude grid matching.

\subsection{Data Quality and Property}

\begin{figure}[tbp]
    \centering
    \includegraphics[width=0.8\linewidth]{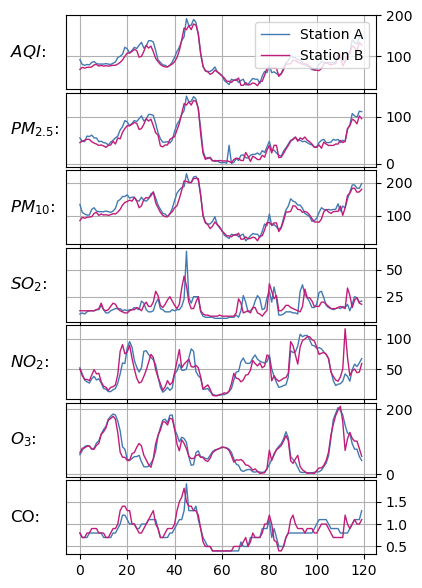}
    \caption{Temporal trends of AQI and pollutant concentrations at station A and station B.}
    \label{fig:data_visualization}
\end{figure}

\begin{figure}[tbp]
    \centering
    \includegraphics[width=0.8\linewidth]{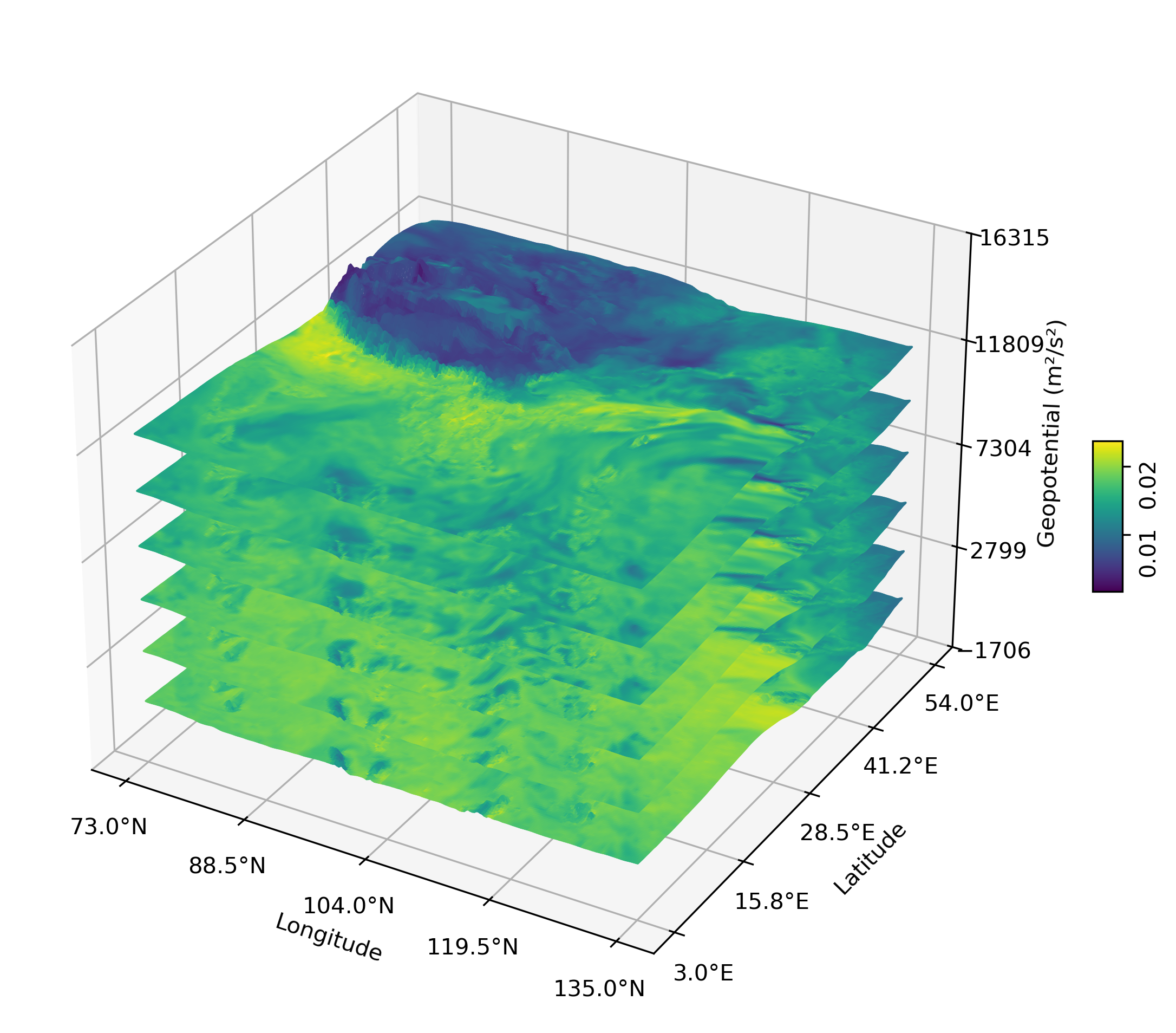}
    \caption{Specific humidity at multiple pressure levels over air quality monitoring stations.}
    \label{fig:multi_level_mete_q}
\end{figure}

The monitoring network covers all provincial capitals and prefecture-level cities across mainland China, with Fig.~\ref{fig:station_distribution} illustrating the station distribution. Compared with existing datasets, our dataset represents the largest scale, encompassing the broadest array of meteorological variables. Detailed statistical characterizations and comparative analyses are systematically presented in Table~\ref{data_tab}, The dataset's unique value lies in its three-way integration of pollution records with both historical meteorological conditions and weather Forecast. Furthermore, the open-access nature of this dataset enhances reproducibility and transparency in environmental research, providing a robust foundation for interdisciplinary collaboration and model validation.

Fig.~\ref{fig:data_visualization} demonstrates significant spatiotemporal coherence in air pollution dynamics: despite their geographical separation exceeding 300 km in Fig.~\ref{fig:station_distribution}, Stations A and B exhibit synchronized temporal pollution concentrations variations. Furthermore, as shown in Fig.~\ref{fig:data_visualization}, significant interactions are observed among the concentrations of the six air pollutants. The concentrations of PM\textsubscript{2.5} and PM\textsubscript{10} at the same location often display similar trends, while NO$_2$ and O$_3$ are often inversely related. Fig.~\ref{fig:multi_level_mete_q} illustrates specific humidity distributions across multiple pressure levels over air quality monitoring stations. The pressure layers correspond to distinct altitudes, and each level exhibits unique humidity distribution characteristics.

\section{Experiments}
To evaluate the performance of the proposed method, we conduct comparative and ablation experiments on ChinaAirNet and KDDCUP-Beijing. The Beijing dataset is evaluated at the city scale, while the ChinaAirNet dataset is evaluated at the national scale. The statistics and comparisons of these two datasets are presented in Table~\ref{data_tab}. To establish an idealized forecasting scenario for methodological validation, we utilize future 48-hour ground truth meteorological measurements from ERA5 as forecast surrogates. This design decision is based on the technical considerations that the operational ECMWF NWP system~\cite{aifsecmwf} and AI weather prediction model~\cite{graphcast, gencast, pangu} achieves high short-term forecasting accuracy (mean absolute error $<2\%$ for temperature and wind speed within 48-hour horizon), making historical observations appropriate proxies for forecast approximation. Each forecast record contains identical variables and spatial resolution as historical meteorological data.

\begin{table*}[tp]
\centering
\caption{Performance Comparison of Different Approaches. A smaller value indicates better performance.}
\begin{tabular}{l|c c c c c c |c c c c c c }
\toprule
\midrule
\multirow{2}{*}{\textbf{Models}} & \multicolumn{6}{c|}{\textbf{ChinaAirNet}} & \multicolumn{6}{c}{\textbf{KDDCUP-Beijing}} \\
\cmidrule(lr){2-7} \cmidrule(lr){8-13}
 & \multicolumn{2}{c}{1-12h} & \multicolumn{2}{c}{13-24h} & \multicolumn{2}{c|}{25-48h} & \multicolumn{2}{c}{1-12h} & \multicolumn{2}{c}{13-24h} & \multicolumn{2}{c}{25-48h} \\
 & MAE & RMSE & MAE & RMSE & MAE & RMSE & MAE & RMSE & MAE & RMSE & MAE & RMSE \\
\midrule
ARIMA~\cite{kumar2010arima} & 32.876 & 50.753 & 33.879 & 51.990 & 34.818 & 53.797 & 30.866 & 39.531 & 44.466 &54.532 & 49.992 & 65.285 \\
HA~\cite{HA} & 29.442 & 44.875 & 30.655 & 46.803 & 32.304 & 49.669  &33.750 & 42.122 & 42.163 & 51.776 & 48.831 & 61.487 \\
\midrule
DCRNN~\cite{li2018dcrnn_traffic} & 19.836 & 30.570 & 25.278 & 37.838 & 29.169 & 43.173 & 26.597 & 36.522 & 36.766 & 49.364 & 42.952 & 56.172 \\
STGCN~\cite{stgcn} & 19.963 & 30.876 & 23.861 & 36.520 & 26.808 & 40.471 & 23.951& 33.362  &34.593 &44.691 & 39.158 & 53.385 \\
ASTGCN~\cite{ASTGCN} & 23.229 & 35.768 & 24.126 & 37.056 & 25.957 & 39.336 & 24.381 & 34.578 & 32.702 & 45.145 & 41.120 & 55.839 \\
GMAN~\cite{zheng2020gman} & 20.732 & 31.291 & 26.057 & 38.564 & 29.725 & 43.674  & 24.036 & 34.044 & 35.747 & 48.024 &  41.754 & 55.914\\
MTGNN~\cite{mtgnn} & 17.976 & 34.552 & 23.432 & 39.909 & 26.828 & 43.479 & 23.979 & 33.774 & 35.919 & 47.974 & 40.780 & 52.902 \\
\midrule
PatchTST~\cite{patchtst} & 17.395 & 28.398 & 22.851 & 35.697 & 26.157 & 40.159 & 22.855 & 32.011 & 34.798 & 43.506 & 39.283 & 49.025 \\
Crossformer~\cite{crossformer} & 16.120 & 27.169 & 22.265 & 35.192 & 26.361 & 40.619 & 21.624 & 31.839 & 33.185 & 42.307 & 40.697 & 52.050 \\
iTransformer~\cite{liu2024itransformer} & 16.065 &25.390  & 21.754   &32.621   &25.051  &36.928   &21.157 &\underline{30.424} & 32.837 &42.113 &\underline{38.566} &\underline{48.589} \\
\midrule
AirFormer~\cite{liang2023airformer} & \underline{15.401} & \underline{24.136} & \underline{21.437} & \underline{31.971} & 25.017 & 36.731 & 21.388 & 31.022 &31.227 &42.752 &38.693 & 48.660  \\
STAFNet~\cite{STAFNet} & 15.896 & 26.938 & 21.855 & 34.764 & 25.684 & 37.864  & 21.540 & 30.945 &32.704 & \underline{41.873} & 39.004 & 48.864  \\
AirPhyNet~\cite{airphynet} & 16.682 & 25.829 & 21.674 & 32.290 & \underline{24.661} & \underline{36.293} & \underline{21.086} & 30.693 & \underline{30.973} & 42.845 &  38.576 & 48.664    \\
\midrule
\textbf{MDSTNet} & \textbf{14.182} & \textbf{22.838} & \textbf{17.933} & \textbf{28.068} & \textbf{19.759} & \textbf{30.724} &\textbf{20.144} & \textbf{29.591} & \textbf{28.413} &\textbf{39.370} & \textbf{35.636} & \textbf{45.380} \\
\midrule
\bottomrule
\end{tabular}
\label{table:comparison}
\end{table*}

\subsection{Implementation Details}
For model parameter setting, we set the encoder layer and decoder layer $L = 3$. The number of DAM tokens in the Spatial Attention, Pollutant Variate Attention, and Meteorological Attention components are fixed at $50, 3, 1$ respectively. The embedding dimension $D$ is set to 256. The model is trained using the Adam optimizer~\cite{kingma2015adam} with a learning rate of $lr = 0.0005$ for 30 epochs, with a batch size of 4. Following previous studies~\cite{STAFNet,han2022semi, hankill2}, the number of historical time steps $T$ and future time steps $\tau$ are set to 24 and 48, respectively. All deep learning models are implemented using PyTorch. All experiments are conducted on a Linux server equipped with a NVIDIA A6000 GPU.

\subsection{Baselines}
We compare the performance of our model with the following three categories and 13 baselines. For a fair comparison, we carefully fine-tuned the hyper-parameters of each baseline on our datasets via grid search. Also, note that all the baselines use the same input features as ours.

\begin{itemize}
\item[$\bullet$] $\textbf{\textit{Classical methods}}$: \\
HA~\cite{HA}: Computes predictions through historical averaging of air quality observations.\\
ARIMA~\cite{kumar2010arima}: Integrates autoregressive and moving average components with differencing for non-stationary time series forecasting.

\item[$\bullet$] \textbf{\textit{STGNN-based models}}: \\
     DCRNN~\cite{li2018dcrnn_traffic}: Combines diffusion convolution with encoder-decoder architecture for spatiotemporal dependency learning.\\
    STGCN~\cite{stgcn}: Designs spatial-temporal synchronous modules using graph convolution and temporal convolution networks. \\
    ASTGCN~\cite{ASTGCN}: Incorporates adaptive spatial-temporal attention for dynamic correlation modeling.\\
    GMAN~\cite{zheng2020gman}: Develops multi-head spatiotemporal attention with gated fusion mechanisms.\\
     MTGNN~\cite{mtgnn}: Proposes mix-hop graph neural networks with expanded inception layers for multi-scale feature extraction.

\item[$\bullet$] \textbf{\textit{Time-series prediction models}}:\\
     PatchTST~\cite{patchtst}: Implements patching strategy to enhance Transformer's local context modeling in time series.\\ 
     Crossformer~\cite{crossformer}: Constructs dimension-scale attention for cross-variable interaction learning.\\
     iTransformer~\cite{liu2024itransformer}: Inverts traditional architecture by treating time points as tokens in attention computation.

\item[$\bullet$] $\textbf{\textit{Air quality prediction models}}$:\\
    AirPhyNet~\cite{airphynet}: Integrates physics-guided neural modules with meteorological constraint learning. \\
    AirFormer~\cite{liang2023airformer}: Leverages spatiotemporal Transformer with air pollution diffusion modeling.\\
    STAFNet~\cite{STAFNet}: Designs adaptive feature fusion modules for spatiotemporal air quality patterns.

\end{itemize}

\begin{table*}[ht]
\centering
\caption{Model Performance Comparison on ChinaAirNet with Different Meteorological Inputs (1-48h Forecast Horizon).}
\begin{tabular}{l | ccc | ccc | ccc}
\midrule
\midrule
\multirow{2}{*}{\textbf{Input Data}} & \multicolumn{3}{c}{\textbf{AirFormer}} & \multicolumn{3}{c}{\textbf{iTransformer}} & \multicolumn{3}{c}{\textbf{MDSTNet}} \\
\cmidrule(lr){2-4} \cmidrule(lr){5-7} \cmidrule(lr){8-10}
 & \textbf{MAE} & \textbf{RMSE} & \textbf{MAE\%} & \textbf{MAE} & \textbf{RMSE} & \textbf{MAE\%} & \textbf{MAE} & \textbf{RMSE} & \textbf{MAE\%} \\
\midrule
AQ-only                                    & 22.657 & 26.763 & -      & 22.817 & 36.199 & -      & 20.623 & 31.540   & - \\
AQ + Surface Mete                          & 22.175 & 26.565 & 2.13\%   & 22.570 & 36.111 & 1.08\%   & 20.399 & 30.835  & 1.09\% \\
AQ + Surface Mete + Forecast               & -      & -      & -      & 22.270 & 35.821 & 2.40\%   & 19.484 & 30.250    & 5.53\%\\
AQ + Multi-pressure-levels Mete            & 21.718 & 32.799 & 4.15\%   & 22.345 & 35.728 & 2.07\%   & 18.869 & 29.275 & 8.51\% \\
AQ + Multi-pressure-levels Mete + Forecast & -      & -      & -      & \textbf{21.980} & \textbf{33.582} & \textbf{3.67\% }  & \textbf{17.908} &\textbf{ 28.272} & \textbf{13.16\%} \\
\midrule \midrule
\end{tabular}
\begin{minipage}{\textwidth}
\vspace{2pt}
\footnotesize 
\emph{Note:} "AQ" refers to air quality monitoring data. "Mete" denotes historical meteorological records. "Forecast" represents weather prediction data (0-48h horizon). All metrics (MAE, RMSE, MAE\%) are calculated for 1-48 hour forecasting horizon. The en dash (–) indicates cases where corresponding experiments were not conducted. MAE\% represents the percentage improvement over baseline (AQ-only).
\end{minipage}
\label{table:input}
\end{table*}

\begin{table*}[ht]
\centering
\caption{Ablation Study on MDSTNet Architecture Variants.}
\begin{tabular}{c| c c c | c c | c c c c c c | c c}
\midrule
\midrule
\multirow{2}{*}{\textbf{Structure }} & \multicolumn{3}{c|}{\textbf{Encoder}} & \multicolumn{2}{c|}{\textbf{Decoder}} & \multicolumn{2}{c}{\textbf{1-12h}} & \multicolumn{2}{c}{\textbf{13-24h}} & \multicolumn{2}{c}{\textbf{25-48h}} & \multicolumn{2}{c}{\textbf{1-48h}} \\
\cmidrule(lr){2-4} \cmidrule(lr){5-6} \cmidrule(lr){7-8} \cmidrule(lr){9-10} \cmidrule(lr){11-12} \cmidrule(lr){13-14}
 & \textbf{Spa} & \textbf{Pva} & \textbf{Mva} & \textbf{Spatial} & \textbf{Variable} & \textbf{MAE} & \textbf{RMSE} & \textbf{MAE} & \textbf{RMSE} & \textbf{MAE} & \textbf{RMSE} & \textbf{MAE} & \textbf{RMSE} \\
\midrule
Decouple Parallel & \Checkmark & \XSolidBrush & \XSolidBrush & \XSolidBrush & \XSolidBrush & 15.717 & 24.600 & 21.271 & 31.648 & 24.868 & 36.351 & 21.681 & 32.865 \\
Decouple Parallel & \XSolidBrush & \Checkmark & \XSolidBrush & \XSolidBrush & \XSolidBrush & 15.491 & 24.702 & 21.685 & 32.396 & 25.096 & 36.758 & 21.842 & 33.263 \\
Decouple Parallel & \XSolidBrush & \XSolidBrush & \Checkmark & \XSolidBrush & \XSolidBrush & 15.329 & 24.122 & 19.893 & 29.931 & 23.205 & 34.089 & 20.408 & 31.083 \\
Decouple Parallel & \XSolidBrush & \Checkmark & \Checkmark & \XSolidBrush & \XSolidBrush & 14.573 & 23.073 & 18.708 & 28.353 & 21.349 & 31.923 & 18.995 & 29.338 \\
Decouple Parallel & \Checkmark & \XSolidBrush & \Checkmark & \XSolidBrush & \XSolidBrush & 14.674 & 23.163 & 18.739 & 28.295 & 21.201 & 31.816 & 18.954 & 29.266 \\
Decouple Parallel & \Checkmark & \Checkmark & \XSolidBrush & \XSolidBrush & \XSolidBrush & 15.377 & 24.247 & 20.488 & 30.674 & 23.313 & 34.516 & 20.623 & 31.540 \\
Decouple Parallel & \Checkmark & \Checkmark & \Checkmark & \XSolidBrush & \Checkmark & 14.632 & 23.123 & 18.596 & 28.339 & 20.490 & 31.011 & 18.552 & 28.840 \\
Decouple Parallel & \Checkmark & \Checkmark & \Checkmark & \Checkmark & \XSolidBrush & 14.349 & 23.003 & \underline{18.131} & \underline{28.171} & 20.272 & 31.337 & 18.256 & 28.665 \\
Integrated & \Checkmark & \Checkmark & \Checkmark & \Checkmark & \Checkmark & 15.521 & 24.064 & 20.068 & 29.858 & 22.700 & 33.298 & 20.247 & 30.622 \\
Decouple Serial & \Checkmark & \Checkmark & \Checkmark & \Checkmark & \Checkmark & \underline{14.324} & \underline{22.956} & 18.182 & 28.214 & \underline{20.060} & \underline{30.979} & \underline{18.157} & \underline{28.471} \\
\midrule
\multicolumn{6}{c|}{\textbf{MDSTNet}} & \textbf{14.182} &\textbf{22.838}  & \textbf{17.933} & \textbf{28.068} & \textbf{19.759} & \textbf{30.724}   & \textbf{17.908} & \textbf{28.272} \\

\midrule \midrule
\end{tabular}
\begin{minipage}{\textwidth}
\vspace{2pt}
\footnotesize 
\emph{Note:} The checkmark and cross symbol indicate whether the corresponding branch is active or blocked. "Structure" denotes whether the model employs an Integrated, serial, or parallel architecture. "Integrated" represents capturing multi-dimensional feature correlations with the unified branch. “Decouple Serial” denotes that distinct branches are structured in a sequential topology.
\end{minipage}
\label{table:structural parallel architecture}
\end{table*}
\subsection{Evaluation Metrics}
Following previous studies~\cite{GC-DCRNN,han2022semi}, we evaluate our method on ChinaAirNet by examining the prediction accuracy of the AQI index which quantitatively and comprehensively describes the air quality situation. For the KDDCUP-Beijing dataset, which lacks AQI records, we evaluate the prediction of PM\textsubscript{2.5} concentrations as defined in prior studies~\cite{Air-DualODE}. We adopt two widely used metrics for evaluation: Mean Absolute Error (MAE) and Root Mean Squared Error (RMSE), which is defined as:

\begin{equation}
\operatorname{MAE }=\frac{1}{\tau N} \sum_{n=1}^N  \left|\hat{\mathbf{X}}_{n,c}^{T+1:T+\tau} -\mathbf{X}_{n,c}^{T+1:T+\tau}\right|,\\
\end{equation}
\begin{equation}
\quad  \operatorname{RMSE } =\sqrt{\frac{1}{\tau N} \sum_{n=1}^N   \left|\hat{\mathbf{X}}_{n,c}^{T+1:T+\tau} -\mathbf{X}_{n,c}^{T+1:T+\tau}\right|^2 }.
\end{equation}

\subsection{Performance Comparison}
In this section, we perform a model comparison in terms of MAE and RMSE. The performance of our method and other prediction baselines is shown in Table~\ref{table:comparison}. The results demonstrate the remarkable effectiveness of the MDSTNet. Our method exhibits significant improvements across all time periods in comparison to all competing baselines on both datasets. Compared with the most competitive method AirFormer, our MDSTNet achieves (7.92\%, 16.35\%, 21.02\%) improvements in terms of MAE for (1-12h, 13-24h, 25-48h) AQI prediction. Moreover, MDSTNet achieves a (4.47\%, 8.27, 7.62\%) performance advantage over AirPhyNet in 48-hour PM\textsubscript{2.5} prediction tasks on city-scale datasets. The multi-pressure-level meteorological features in ChinaAirNet allow MDSTNet to achieve enhanced performance gains.
From Table~\ref{table:comparison} we can also observe that the deep learning models significantly outperform the statistical approaches. STGNN-based models, with their ability to learn complex relationships, outperform classical methods. 
The PatchTST approach generates predictions in a channel-independent form, which is incapable of capturing the intricate interdependencies. Both the iTransformer and Crossformer models employ channel attention, yet neither of them treats exogenous variables separately. Furthermore, Crossformer only employs the router mechanism on channel attention to reduce complexity, whereas the DAM proposed in MDSTNet is utilized in multiple correlation modeling and also equipment support for exogenous variables.
Moreover, models (AirFormer, AirPhyNet) specifically designed for air quality prediction that incorporate domain knowledge demonstrate superior predictive accuracy. Our MDSTNet clearly outperforms AirFormer which is also a transformer model for air quality prediction. 
Overall, MDSTNet exhibits superior performance across diverse dataset scales and pollutant prediction scenarios, with particularly pronounced advantages on larger-scale datasets and extended forecasting horizons. This substantiates the effectiveness of MDSTNet's encoder-decoder architecture in systematically leveraging comprehensive meteorological data and multi-step weather forecast integration for enhanced prediction accuracy.

\subsection{Ablation Study}
In this section, we conduct an ablation study on ChinaAirNet to assess the effectiveness of our proposed modules.

\subsubsection{Effect of multi-pressure-level meteorological data and Weather forecast}
To investigate the impact of multi-pressure-level meteorological data and weather forecasts on air quality prediction, we conducted ablation experiments using AirFormer, iTransformer, and our proposed MDSTNet to comprehensively analyze the influence of different feature data and compare the models' utilization of information. 
The results shown in Table~\ref{table:input} clearly demonstrate that atmospheric conditions are crucial for air quality prediction. Incorporating historical meteorological data improves prediction accuracy for AirFormer, iTransformer, and MDSTNet, with more significant gains observed when using more comprehensive multi-pressure-level meteorological data, which proves that multi-pressure-level meteorological data contains additional information beneficial to air quality prediction tasks compared to surface-level meteorological features. Furthermore, iTransformer and MDSTNet can leverage weather forecasts to further reduce prediction errors, while AirFormer is structurally constrained from incorporating multi-step weather forecasts, demonstrating that accurate weather forecasts can guide models to make more precise air quality predictions. Notably, MDSTNet exhibits more substantial improvements when atmospheric conditions are included, indicating that its architecture is better suited for modeling the spatiotemporal correlations among different features.

\subsubsection{Effect of MDSTNet architecture}
To investigate the effects of the parallel encoder-decoder architecture in MDSTNet and the contributions of individual branches in encoder and decoder to final predictions, we conducted detailed ablation experiments comparing multiple MDSTNet variants and the results shown in Table~\ref{table:structural parallel architecture}. 
The results demonstrate that decoupling the modeling of spatial, variable, and meteorological correlations significantly enhances prediction accuracy, with the parallel structure outperforming serial architectures. We believe this is because the decoupled parallel structure allows each branch to learn different attention dependency patterns when modeling corresponding correlation, thereby avoiding interference from distributional biases among different correlations. The encoder-decoder architecture further improves performance by separately modeling historical spatiotemporal patterns and future forecast information, effectively establishing dynamic temporal dependencies across different temporal features. This further confirms that weather forecast information can serve as a prompt to guide precise air quality predictions. Additionally, the encoder components Spa, Pva, and Mva collectively contribute to performance improvement, with Mva demonstrating the most substantial gains, which highlight the critical importance of meteorological information in prediction. The decoder's attention mechanisms operating through both spatial and variable dimensions are equally essential, indicating that weather forecast guidance through both spatial and variable perspectives.

\begin{figure}[ht]   
  \centering            
  \subfloat[ Prediction Accuracy ]   
  {
      \label{spa DAM tokens}\includegraphics[width=0.34\textwidth]{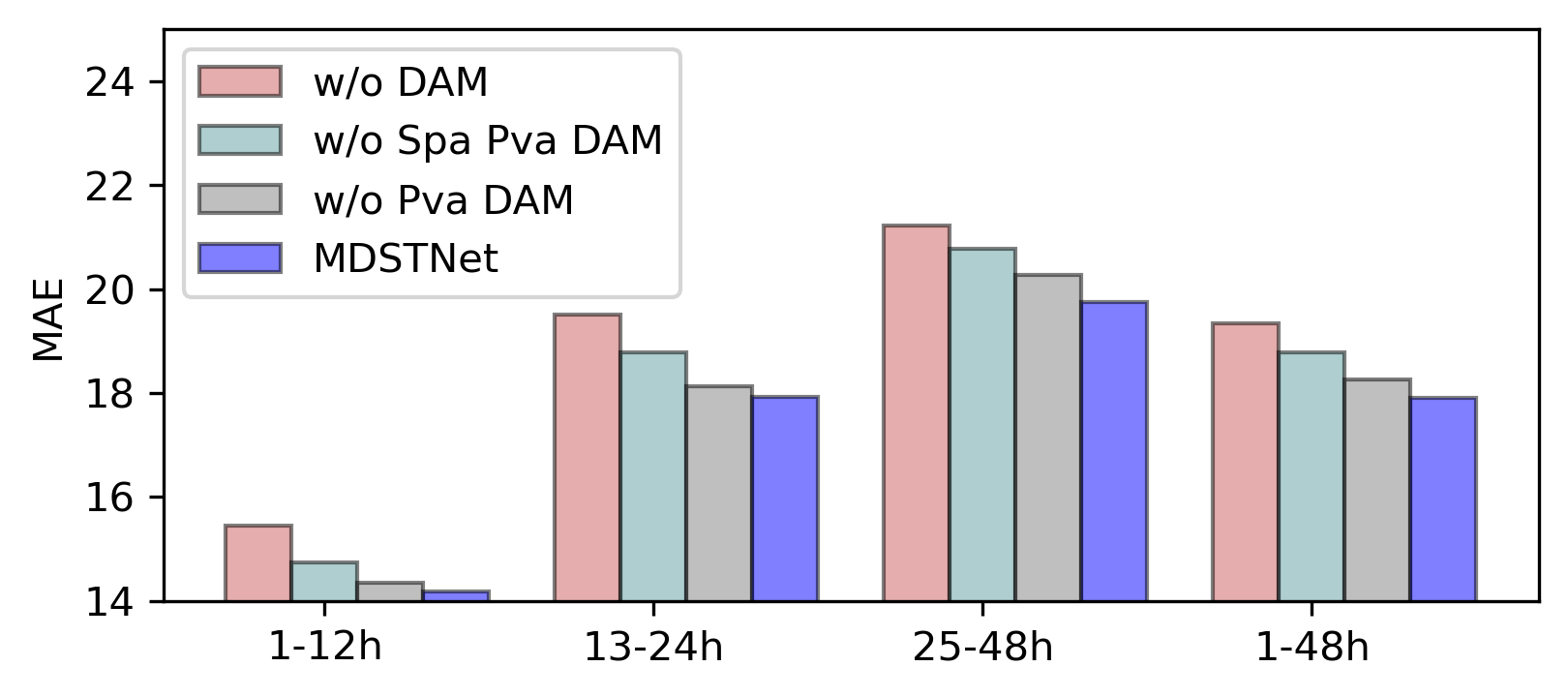}
     
  }
  \subfloat[ Training Time ]
  {
      \label{pva DAM tokens}\includegraphics[width=0.14\textwidth]{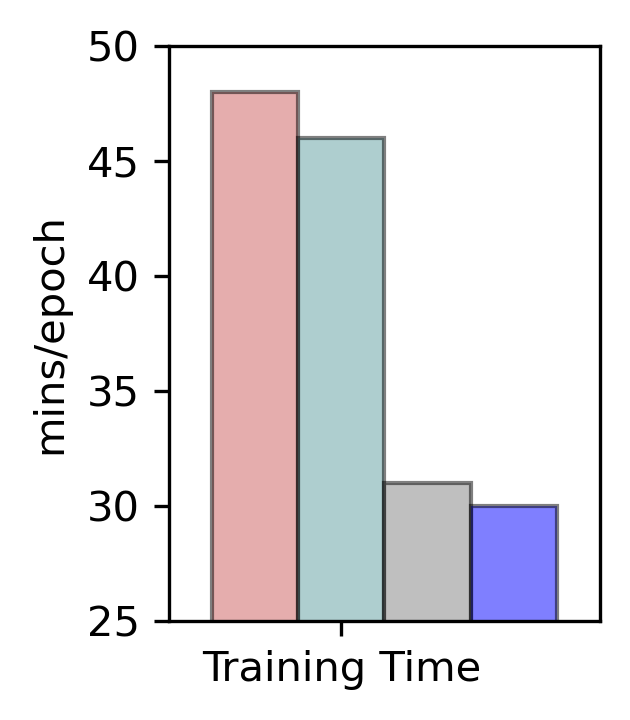}
  }
  \caption{ Effect of the Distillation Attention Module.}    
  \label{Effect of DAM}           
\end{figure}

\subsubsection{Effect of DAM}
To examine the efficacy of DAM, we compare our model with its variants: \textbf{w/o Spa/Pva/Mva DAM}: we gradually replace the DAM in Spa, Pva and Mva with standard MSA. The results are shown in Fig.~\ref{Effect of DAM}. We observe that the use of DAM enhances predictive accuracy in both spatial and variate correlations while simultaneously reducing computational costs. Compared to the \textbf{w/o DAM}, MDSTNet achieves a 37.5\% reduction in training time and a 7.4\% decrease in the average MAE across 1-48h. \textbf{This demonstrates that DAM effectively removes data redundancy and improves the signal-to-noise ratio in spatiotemporal representations.}
\begin{figure}[t]
\centering
\includegraphics[width=0.95\columnwidth]{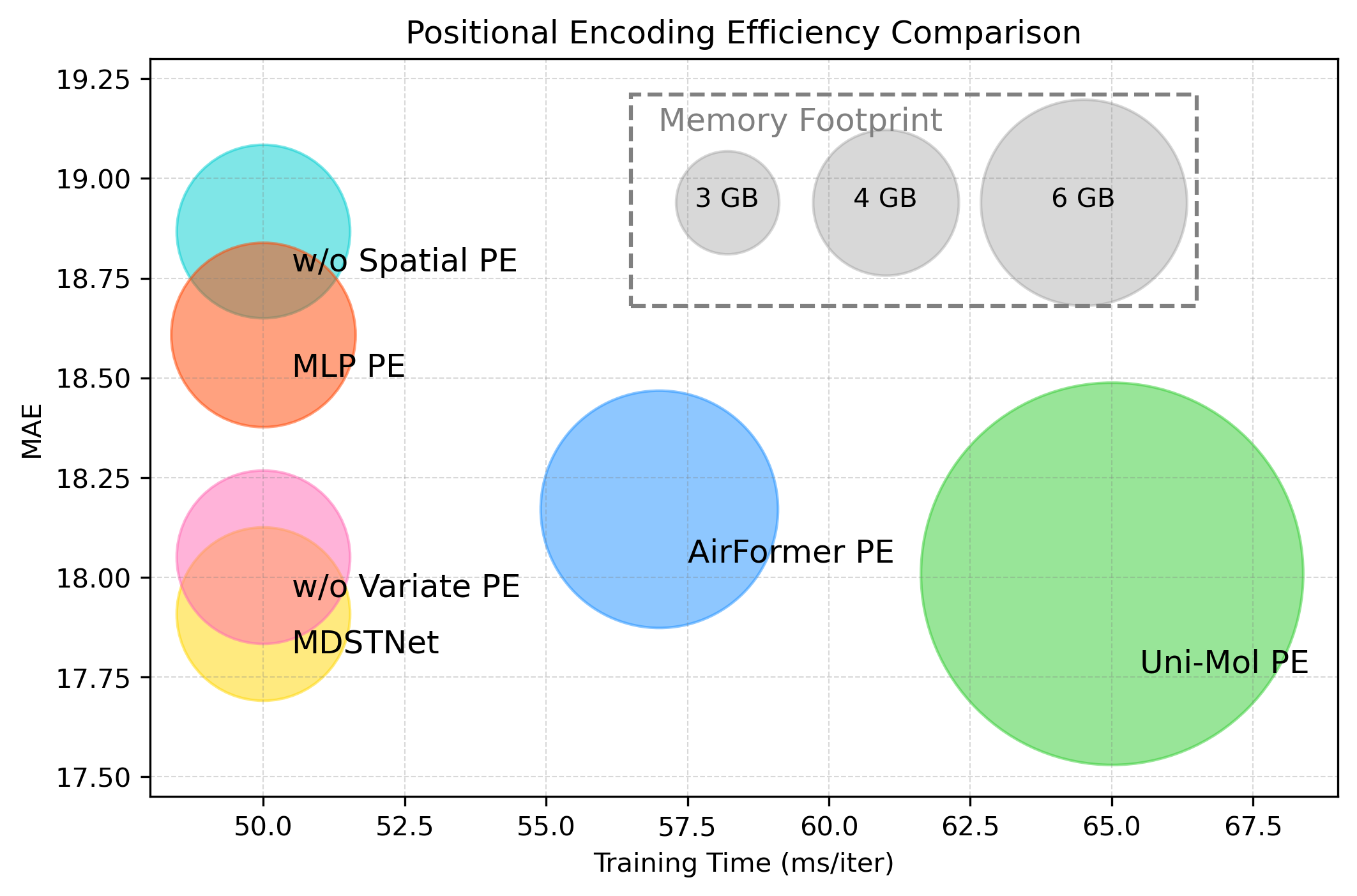} 
\caption{Effect of Spatial and Variate Positional Encoding. The vertical and horizontal axes represent MAE and the training time. The circle's size corresponds to the memory footprint.}
\label{PE}
\end{figure}

\subsubsection{Effect of Spatial and Variate Positional Encoding}
To examine the efficacy of Spatial and Variate Positional Encoding, we compare the forecasting performance, training speed, and memory footprint of the following variants: a) \textbf{w/o Spatial PE}: we remove the Spatial Positional Encoding in embedding lay. b) \textbf{w/o Variate PE}: we remove the Variate Positional Encoding in the embedding lay. c) \textbf{MLP, AirFormer, and Uni-Mol PE} : we replace the MDSTNet Spatial Positional Encoding by MLP, AirFormer~\cite{liang2023airformer}, and Uni-Mol~\cite{Uni-Mol} PE. The results are shown in Fig.~\ref{PE}. Primarily, all Positional Encoding methods significantly outperform w/o Spatial PE, underscoring the necessity of incorporating spatial information in transformers. it is evident that the Spatial PE proposed in this paper achieves performance comparable to that of Uni-Mol PE, while introducing minimal additional computational and memory overhead. Furthermore, the inclusion of Variate PE enhances the model's ability to capture variate correlations.


\subsection{Hyperparameter Study}

We evaluate the hyperparameter sensitivity of MDSTNet on ChinaAirNet to the following factors: the input length $T$, the number of encoder and decoder layers $L$, the embedding dimension $D$ and the number of DAM tokens in the Spa, Pva, and Mva. 

\begin{figure}[tp]   
  \centering            
  \subfloat[Input Length $T$]
  {
      \label{T}\includegraphics[width=0.24\textwidth]{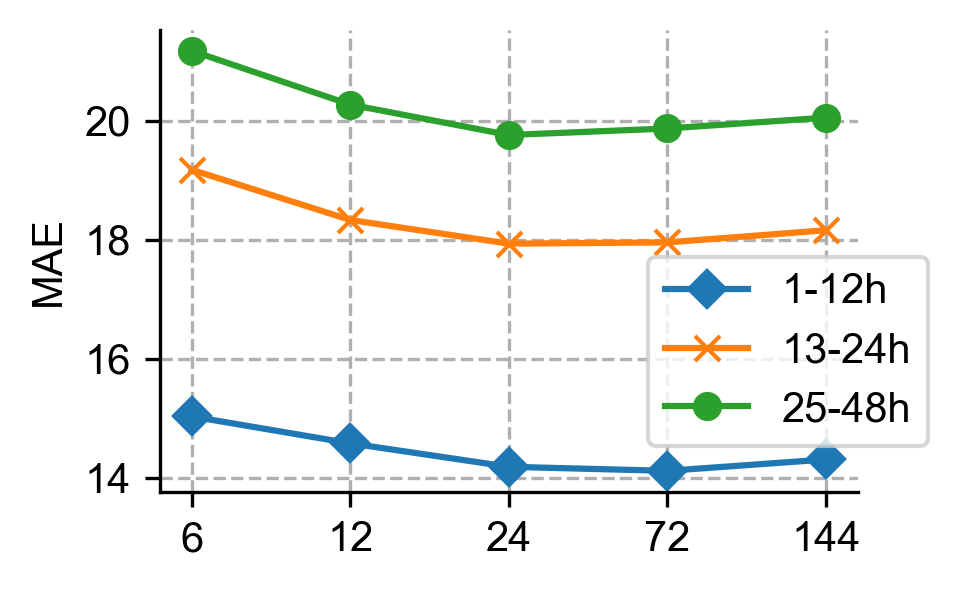}
  }
  \subfloat[ Encoder and Decoder layers $L$]
  {
      \label{L}\includegraphics[width=0.24\textwidth]{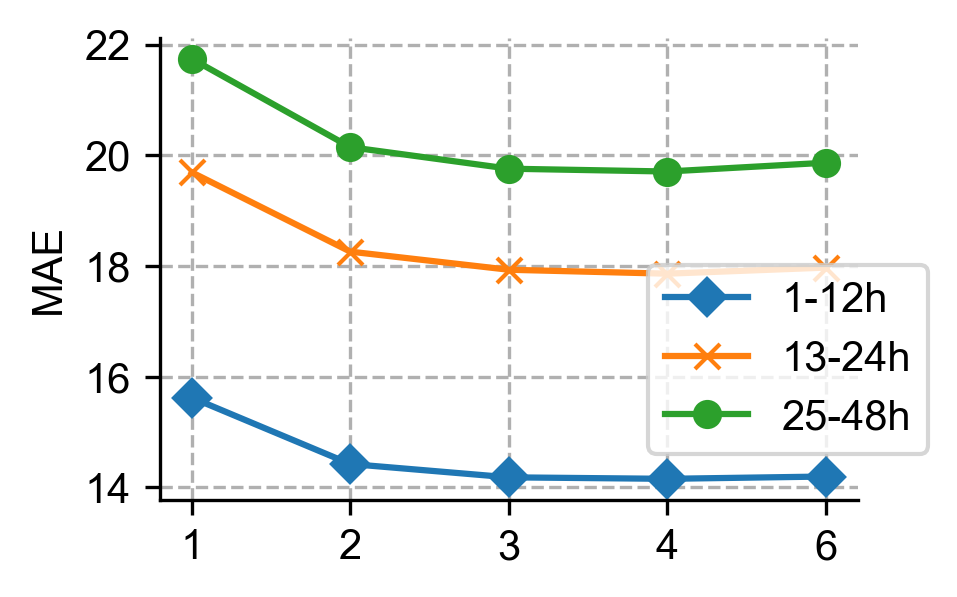}
  }\\
  \subfloat[Embedding Dimension $D$]
  {
      \label{D}\includegraphics[width=0.23\textwidth]{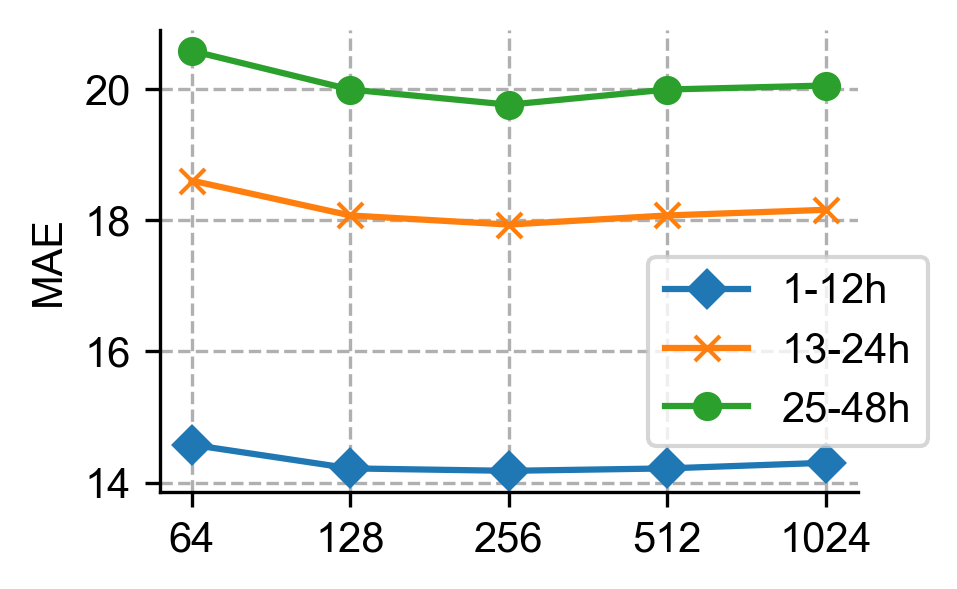}
  }
  \subfloat[Spa DAM tokens]   
  {
      \label{spa DAM tokens}\includegraphics[width=0.23\textwidth]{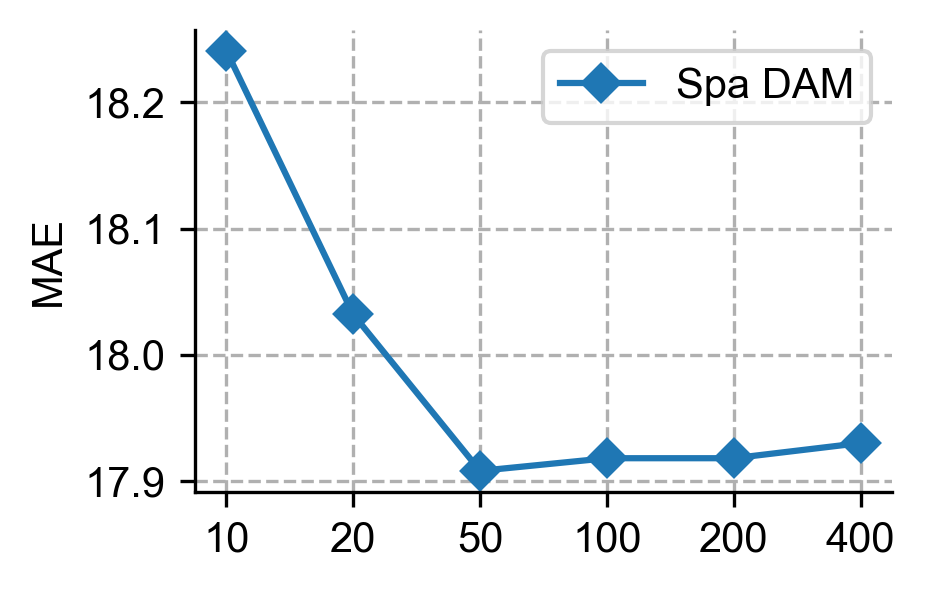}
  }\\
  
    \subfloat[ Pva DAM tokens ]
  {
      \label{pva DAM tokens}\includegraphics[width=0.24\textwidth]{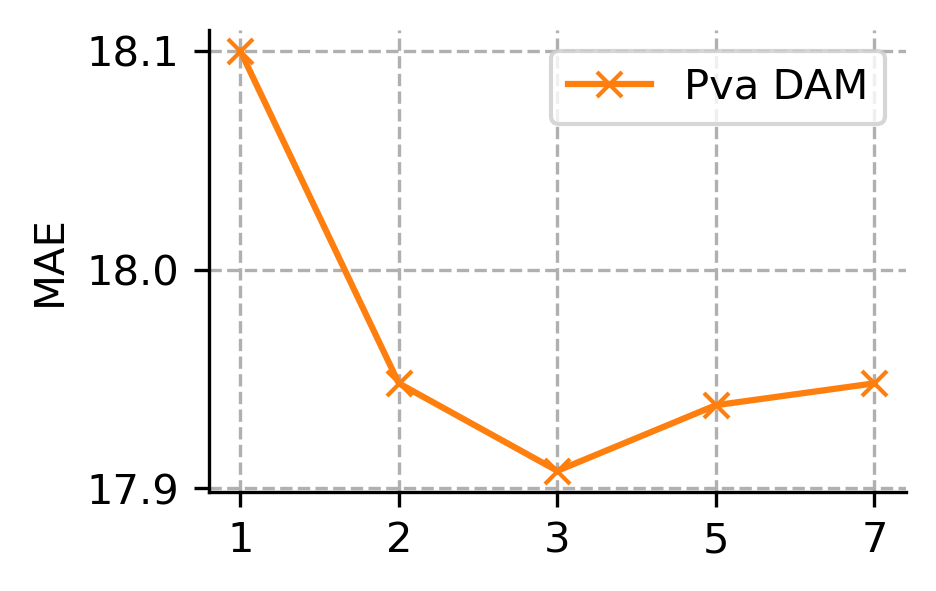}
  }
  \subfloat[ Mva DAM tokens ]
  {
      \label{mva DAM tokens}\includegraphics[width=0.24\textwidth]{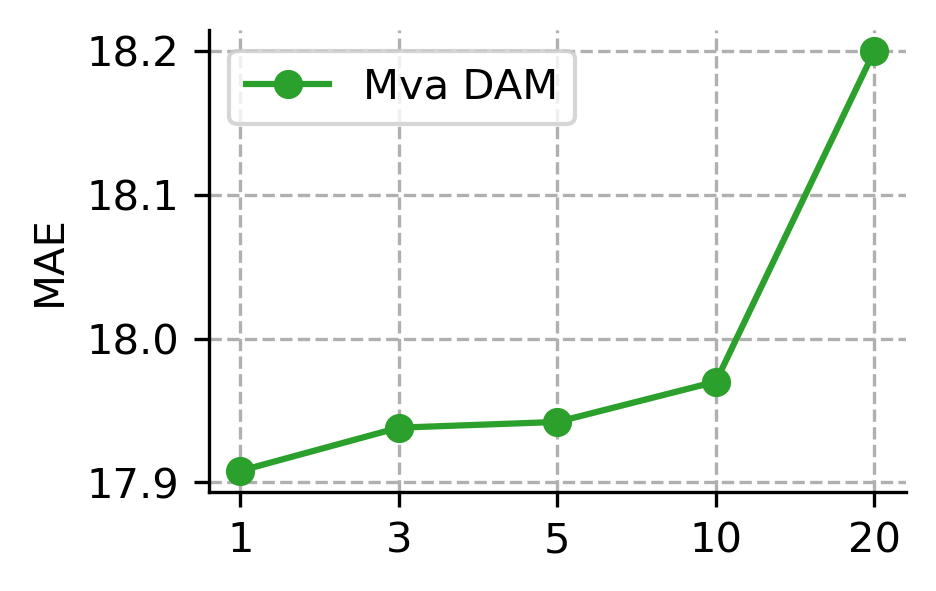}
  }
  \caption{Hyperparameter Study on the ChainAirNet Dataset.}    
  \label{DAM token number}            
\end{figure}

We first vary the input length $T$ from 6 to 144 time steps, with the results presented in Fig.~\ref{T}. Our observations reveal progressive performance improvements as $T$ increases from 6 to 24, followed by marginal degradation when extending the $T$ beyond 24 to 144 time steps. This likely results from two factors: excessive $T$ introduces redundant temporal information that amplifies noise interference in forecasting, while conventional series embedding architectures inherently limit the model's ability to effectively capture long-term temporal dependencies.

We then investigate the impact of encoder and decoder layer counts by varying $L$ from 1 to 6, with results shown in Fig.~\ref{L}. The model performance progressively improves as $L$ increases from 1 to 3 layers, but plateaus when further expanding the architecture to 6 layers. Excessive network depth may cause feature over-smoothing and computational inefficiency, optimal model performance emerges at 3-4 layers.

Experiments investigating the embedding dimension $D$, as shown in Fig.~\ref{D}, demonstrate that increased hidden dimension size yields marginal performance improvements in MDSTNet. This suggests that dimensional expansion provides limited returns beyond optimal configuration thresholds.
 
Fig.~\ref{spa DAM tokens} shows the results that Spa with 10-400 DAM tokens will have a better performance. For Pva DAM in Fig.~\ref{pva DAM tokens}, setting the number of tokens to 3 yields the best performance, which may suggest the presence of three primary patterns within the pollutant variates. In contrast, for meteorological variates in Fig.~\ref{mva DAM tokens}, a single token appears sufficient to encapsulate the meteorological features necessary for accurate prediction. 

\begin{figure*}[htp]
\centering
\includegraphics[width=2\columnwidth]{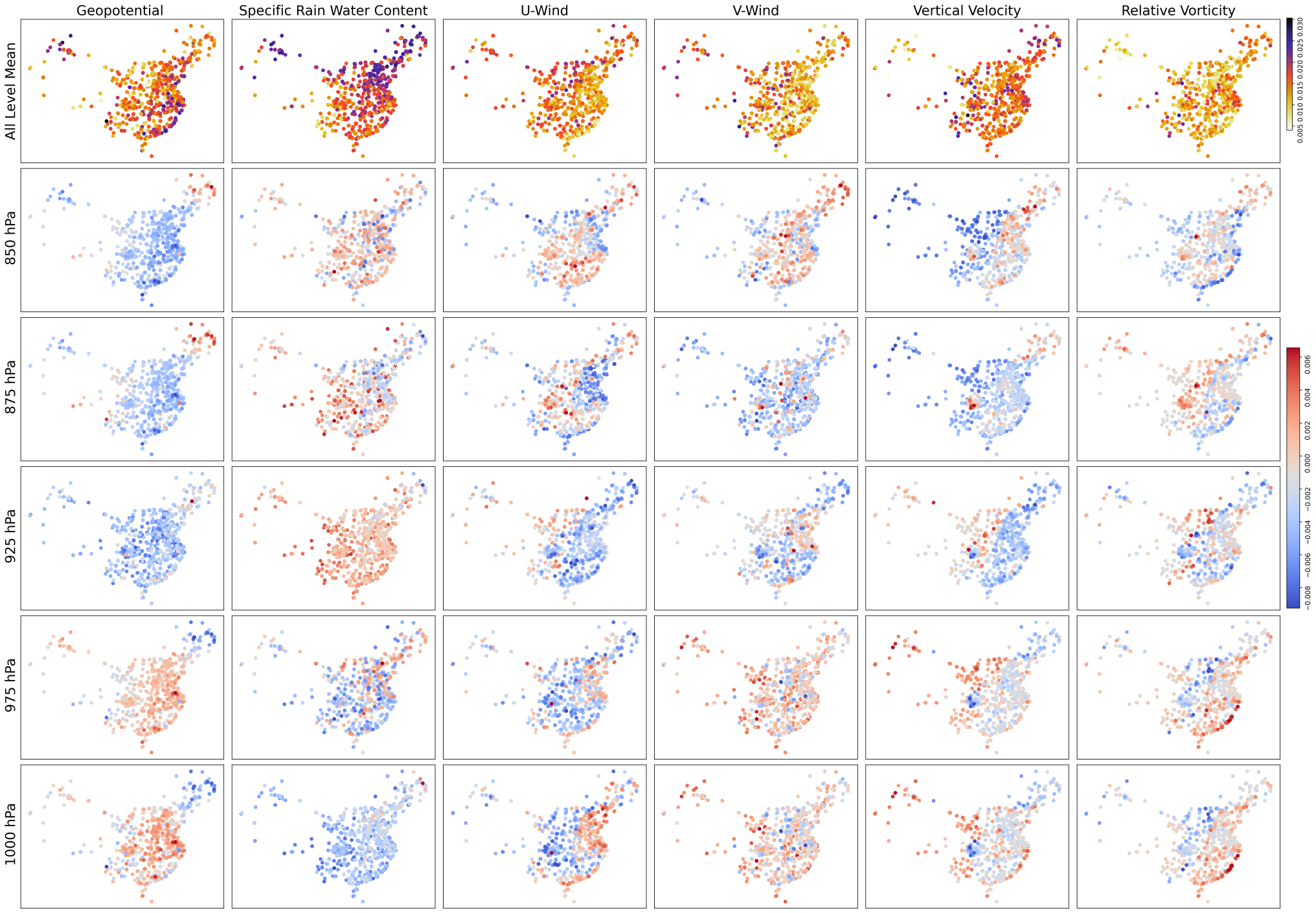} 
\caption{Visualization of Meteorological Attention Distributions at Different Pressure Levels. 
}
\label{multy_level_attention}
\end{figure*}

\subsection{Visualization}

In this section, we qualitatively analyze how our model captures multi-pressure-level meteorological correlations for air quality prediction through visualizations. Due to spatial constraints, Fig.~\ref{multy_level_attention} visualizes attention distributions for a subset of meteorological variables in representative pressure levels. Each column represents attention scores of different meteorological variables at monitoring stations. In the pictures of the first row, color intensity represents the magnitude of averaged attention scores aggregated from all pressure levels for respective meteorological variables, with darker hues indicating stronger variable-specific focus at corresponding stations. To clarify distinctions between pressure levels, subsequent rows illustrate deviations of pressure level-specific attention scores relative to the multi-level mean. Blue dots mark attention reduction relative to the averaged distribution, while red dots signify attention enhancement compared to the multi-level mean. 

The first-row pictures demonstrate that Specific Rain Water Content, Geopotential, Wind Speed, and Vertical Velocity constitute the predominant meteorological variables prioritized across most monitoring stations. Specific Rain Water Content emerges as the dominant focus for northern stations, while southeastern stations exhibit heightened attention to Geopotential, Wind Speed, and Vertical Velocity. This spatial dichotomy suggests regionally differentiated meteorological impacts, where precipitation variables likely play a more critical role in air quality within arid northern regions, whereas atmospheric circulation parameters appear more relevant to pollution dispersion in humid southeastern areas. 

As shown in the last five rows, the attention distributions of different meteorological variables vary across different pressure levels. As exemplified in the first column, most stations demonstrate heightened attention to geopotential height features between 1000 hPa and 975 hPa, whereas the second column shows predominant attention allocation to specific rainwater content characteristics across the 925 hPa to 850 hPa pressure levels. Notably, geospatial variations in vertical attention patterns emerge for identical meteorological variables: stations east of the Taihang Mountains exhibit inverted U-wind speed attention patterns compared to western counterparts. Similarly, stations in the Sichuan Basin display contrasting vertical velocity patterns relative to extra-basin western stations, and coastal stations in southeastern China form distinct spatial clusters in relative vorticity attention. These observations demonstrate the model's capability to capture complex spatiotemporal correlations between multilevel meteorological features across diverse geographical contexts, thereby underscoring the necessity of incorporating multilayer atmospheric data for comprehensive spatiotemporal dependency modeling.

\section{Conclusion and Future Work}

In this paper, We propose a novel framework MDSTNet that pioneers the systematic integration of multi-pressure-level meteorological data and weather forecasts for modeling atmospheric dynamics in air quality prediction. Complementing this architecture, we construct ChinaAirNet—the first nationwide dataset unifying air quality records with multi-pressure-level meteorological observations and forecasts. Comparative experiments demonstrate state-of-the-art performance and reduce the short-range prediction errors by 17.54\% compared to prior methods. The framework's effectiveness is further validated through comprehensive ablation studies and interpretability analysis.


\end{document}